\documentclass{svproc}

\usepackage{url}

\usepackage{amsmath,amssymb}
\usepackage{bm}
\usepackage[numbers,sort&compress]{natbib}
\usepackage{algorithm}
\usepackage{algpseudocode}
\usepackage{booktabs}
\usepackage{multirow}
\usepackage{graphicx}
\usepackage{tikz}
\usetikzlibrary{positioning,calc,arrows.meta,fit,backgrounds}
\usepackage[hidelinks]{hyperref}
\usepackage{soul}
\usepackage{microtype} 

\newcommand{\PR}{\mathbb{P}}
\newcommand{\Map}{\mathcal{M}}
\newcommand{\Geom}{\mathcal{G}}
\newcommand{\Pred}{\Phi}
\newcommand{\Ev}{\mathcal{E}}
\newcommand{\Upd}{\bm{U}}
\newcommand{\Retr}{\bm{R}}
\newcommand{\CRgm}{\mathcal{R}}
\newcommand{\Belief}{\mathcal{B}}
\newcommand{\Conf}{c}
\newcommand{\CalLR}{\hat{\Lambda}}
\newcommand{\Prov}{\mathcal{P}}
\definecolor{BrendanColor}{RGB}{200, 50, 150}

\newif\ifarxiv
\arxivfalse
\newcommand{\arxivonly}[1]{\ifarxiv #1\fi}

\begin{document}
\mainmatter

\title{Belief Consistency Between Foundation-Model Evidence and Geometric Perception in Persistent Robotic Maps}

\titlerunning{Belief Consistency in FM-Augmented Robotic Maps}

\author{Christoffer Heckman\inst{1} \and Harel Biggie\inst{2} \and Brendan Crowe\inst{1} \and Nicholas Roy\inst{2}}

\authorrunning{Heckman, Biggie, Crowe, Roy}


\institute{Department of Computer Science, University of Colorado, Boulder \and Computer Science and Artificial Intelligence Lab, Massachusetts Institute of Technology}

\maketitle

\begin{abstract}

Persistent maps used by autonomous robots increasingly fuse a geometric perception stack whose assertions are well-characterized with a foundation-model channel that produces semantic claims without calibrated reliability about the same scene. Contemporary mapping systems integrate the two channels by treating the foundation-model channel as an additional voter into a per-element posterior, uncalibrated for its own per-class reliability and without machinery to flag when the two channels contradict each other at a given moment. We propose an update operator with two cooperating mechanisms: a per-class calibrated commit gate, and a per-event conflict-drop window that refuses to commit foundation-model claims contradicted by the geometric channel at the moment of the claim. We evaluate on KITTI-360 and ScanNet, with an oracle geometric channel (panoptic ground truth) and an off-the-shelf online semantic segmenter (Mask2Former) to demonstrate real-world performance. The operator produces substantially more accurate committed maps (KITTI \texttt{is\_car} commit precision $99.7\%$ vs.\ $43.9\%$ for the calibration-only operator; mean per-class IoU $0.522$ vs.\ $0.180$), retains more compositional true positives at higher precision than a monolithic compositional VLM prompt. The framework operates at deployment quality across both oracle and off-the-shelf-segmenter geometric channels, and is invariant under foundation-model substitution.
\end{abstract}


\section{Introduction}
\label{sec:intro}

A mobile robot operating in an unstructured environment routinely maintains a persistent map built from two evidence channels with fundamentally different properties: a foundation model that detects and names discrete entities such as objects and semantic segments (a chair, a bookshelf, a parked car) and a geometric perception model that measures these entities (mesh surfaces, occupancy, per-pixel labels).
Contemporary mapping systems \cite{gu2024conceptgraphs,jatavallabhula2023conceptfusion,kerr2023lerf,peng2023openscene,rosinol2021kimera,hughes2022hydra,schmid2024khronos,maggio2024clio,gorlo2026daaam} fuse the two channels by treating the foundation-model channel as an additional voter into a per-element posterior, typically through a windowed majority filter or per-channel recursive Bayesian aggregation.
The voter is uncalibrated for its own per-class reliability, and the fusion has no mechanism to ask whether the claim the foundation model is asserting at this moment is contradicted by what the robot is sensing at this moment.

\begin{figure}[!htbp]
\centering
\includegraphics[width=\linewidth]{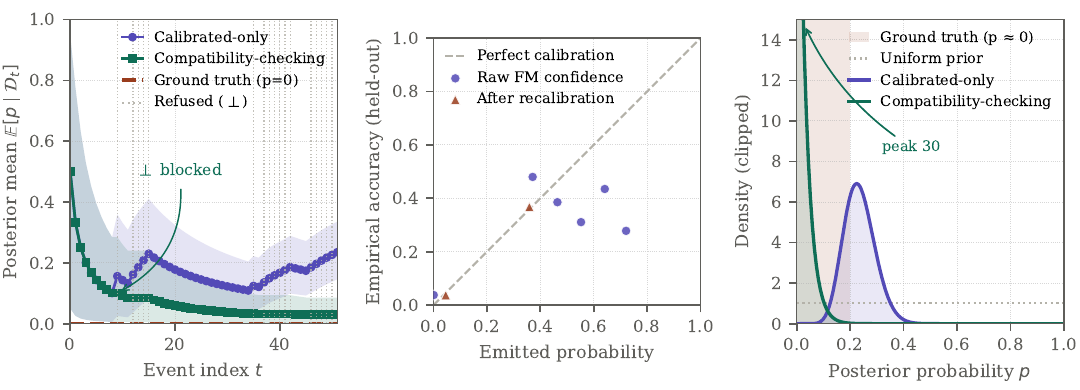}
\caption{Failure modes and their detection on a single $51$-event trace ($18$ geometric-channel contradictions). \emph{(a)} Per-element posterior mean and $90\%$ CI: the calibration-only operator absorbs every event and drifts away from the correct estimate; the compatibility-checking operator refuses contradicting events and holds near ground truth. \emph{(b)} Reliability bins from a single $50/50$ held-out test split ($n=3699$): raw foundation-model confidence (purple) vs.\ conformally calibrated probability (coral). Isotonic regression maps continuous raw confidence to a small number of unique calibrated values, hence fewer coral bins; raw ECE $0.16 \to 0.01$ after calibration. \emph{(c)} Final Beta posteriors: calibration-only mass sits wrong (mean $0.24$); compatibility-checking mass sits at truth (mean $0.03$).}
\label{fig:failure_modes_traces}
\end{figure}

This produces two symptoms that survive into the final map.
The first is \emph{population-level miscalibration:} a foundation model overconfident on a class stays overconfident in the fused map, with the bias compounding under compositional intersection.
The second is \emph{per-event silent contradiction:} a foundation-model claim at a single moment can contradict what the geometric channel reports about the same element at that moment, and the belief state has no mechanism to flag it.
Both produce silently wrong beliefs in the map consumed by downstream components such as a planner or pose estimator.
Geometric perception comes with analytic noise models from characterized sensors and pose-graph consistency conditions formalized in the SLAM literature \cite{lu1997globally}; the foundation-model channel inherits no analogous foundations, and deployed systems carry the gap as a blind spot in their perception stack.
Figure~\ref{fig:failure_modes_traces} shows the two symptoms in a single per-element trace: the foundation-model channel repeatedly asserts a predicate the geometric channel contradicts, the calibration-only operator absorbs every assertion and drifts to a confidently wrong belief, and the per-event compatibility check refuses each contradicting assertion and preserves a belief consistent with the geometric channel.

This work proposes an update operator that addresses both symptoms at the layer where evidence meets belief, by combining two cooperating mechanisms.
First, a \emph{per-class calibrated commit gate} recalibrates the foundation model's confidences for the specific scenario against the geometric channel's verifiable assertions, so that commits reflect the model's empirical reliability on the asserted class rather than its raw output; this addresses the population-level miscalibration.
Second, a \emph{per-event conflict-drop window} emits an explicit inconsistency signal and refuses to commit when the two channels disagree at the same map element at the same moment, with later corroboration within a fixed window resolving the disagreement but not retroactively committing the disagreed-on event; this addresses the per-event silent contradiction.
The two mechanisms are complementary: calibration handles the drift that accumulates when many noisy-but-plausible claims are integrated, and conflict-drop handles the single detection that is obviously wrong given what the robot is sensing right now.
Together they yield a substantially more accurate committed map (mean per-class IoU $0.522$ vs.\ $0.180$ for the calibration-only operator on KITTI), compositional claims retained at higher precision, deployment-quality behavior under an off-the-shelf semantic segmenter, and a calibration mechanism invariant to foundation-model substitution (Brier shifts $\leq 0.004$ under LLaVA-$1.6 \to$ Qwen3-VL swap).

\section{Related Work}
\label{sec:related}

\paragraph{Probabilistic semantic mapping with foundation-model evidence.}
Persistent semantic maps have been built around recursive Bayesian fusion of per-element class posteriors for nearly a decade, with surfel, voxel, panoptic, LiDAR, kernel-Bayesian, and Dirichlet-categorical variants \cite{mccormac2017semanticfusion,sunderhauf2017meaningful,narita2019panopticfusion,chen2019sumapp,grinvald2019volumetric,rosinol2020kimera,doherty2017bayesian,doherty2019learning,gan2020bayesian,wilson2024convbki,asgharivaskasi2021active,asgharivaskasi2023semantic,hornung2013octomap}.
Recent foundation-model systems \cite{gu2024conceptgraphs,jatavallabhula2023conceptfusion,kerr2023lerf,peng2023openscene,hughes2022hydra,hughes2024foundations,bavle2025sgraphs2,li2025hierslam,schmid2024khronos,maggio2024clio,gorlo2026daaam,maggio2025bayesianfields} attach VLM features or per-channel Bayesian aggregation to scene-graph or Gaussian-splat substrates.
Across this lineage the semantic channel is treated as the authoritative truth signal whose noise is reduced by averaging or per-channel Bayesian aggregation; none calibrates the channel's reliability against an independent geometric verifier or flags per-event disagreement at the moment of an event.\arxivonly{ Extended discussion of each subsystem in \S\ref{sec:supp_related}.}
Our framework specializes the per-element Bayesian posterior the lineage provides while disciplining the trust assumption it leaves implicit.

\paragraph{Belief revision and conformal prediction.}
The Alchourr\'on-G\"ardenfors-Makinson postulates \cite{alchourron1985logic,gardenfors1988knowledge} and cognitive-robotics extensions \cite{reiter2001knowledge,lakemeyer2007cognitive} formalize qualitative belief revision; conformal prediction \cite{vovk2005algorithmic,angelopoulos2023gentle,gibbs2021adaptive} supplies distribution-free coverage guarantees, with recent extensions to LLM outputs \cite{quach2024conformal} and robot planning \cite{ren2023robots}.
We compose the two threads by applying conformal recalibration to foundation-model evidence within the persistent map's update operator, using the robot's own geometric perception as the calibration source rather than an offline labeled dataset.\arxivonly{ Random-finite-set and imprecise-probability machinery \cite{mahler2007statistical,walley1991statistical} that informs the formulation but is not operationalized here is surveyed in \S\ref{sec:supp_related_fusion}.}

\section{Problem Statement}
\label{sec:problem}

\paragraph{Preliminaries.}
A \emph{predicate} is a property asserted or denied of a map element (``is a chair,'' ``is traversable on foot,'' ``has been inspected'').
Each element-predicate pair carries a belief $\mathrm{Beta}(\alpha, \beta)$: mean $\alpha/(\alpha+\beta)$ is the expected truth probability, $\alpha+\beta$ the effective sample size, and the distribution is closed under Bayesian updates from Bernoulli observations, allowing for a two-number summary of the element-predicate marginal distribution.
Conformal prediction \cite{vovk2005algorithmic,angelopoulos2023gentle} produces prediction sets with finite-sample distribution-free coverage guarantees, with adaptive variants \cite{gibbs2021adaptive} handling distribution shift.
We use ``conformal recalibration'' as shorthand for applying the same nonconformity-quantile machinery to recalibrate the foundation-model channel's likelihood rather than to emit a prediction set directly; both are marginal-coverage guarantees on exchangeable data.

\subsection{Two Evidence Channels With Distinct Epistemic Properties}
\label{sec:asymmetry}

A robot maintains a persistent map populated by two evidence channels with different characterizations.
The \emph{geometric channel} produces assertions over two predicate classes that the operator treats as authoritative for calibration purposes, even though they carry their own noise.
Pure geometric predicates (spatial adjacency, traversability for a vehicle of given dimensions, occupancy) come from sensors with known or learnable noise models: range sensors, cameras, IMUs, and the SLAM front-ends and back-ends that fuse them, with well-understood consistency conditions.
Class-presence predicates (object of a known class at sensor resolution) come from a verification source paired to the geometric structure: hand-curated annotations, calibrated semantic segmenters with characterized error rates, or human-verified panoptic ground truth from a dataset's annotation pipeline.
The asymmetry the framework exploits is characterized-versus-uncharacterized rather than sensor-versus-model: the geometric channel's noise is analytic and bounded enough for the operator to use it as a calibration reference, while the foundation-model channel's noise is not.

The \emph{foundation-model evidence channel} produces semantic claims via prompted vision-language and language models.
The claims span a far broader vocabulary than the geometric channel can verify, but the channel itself has no analytic noise model: foundation models hallucinate, exhibit confident incorrectness, and produce systematically different errors under different prompting.
The reported confidence is at best an uncalibrated proxy for actual reliability \cite{ren2023robots}.

The two channels are not symmetric: the geometric channel is restricted but with a well-characterized error distribution; the foundation-model channel is expansive but without characterization of errors.
Integration requires machinery for translating between these regimes.
The absence of such machinery is what allows two specific failure modes to go undetected: population-level miscalibration of foundation-model confidences against the geometric channel's outcomes, and per-event cross-channel disagreement that is silently merged into the belief state.

\begin{figure}[!htbp]
\centering
\resizebox{\textwidth}{!}{%
\begin{tikzpicture}[
  >=stealth,
  font=\small,
  box/.style={draw, rounded corners=2pt, align=center, inner sep=2.5pt},
  vlmbox/.style={box, fill=blue!10, minimum height=10mm, minimum width=20mm, font=\footnotesize},
  geombox/.style={box, fill=green!12, minimum height=10mm, minimum width=20mm, font=\footnotesize},
  calbox/.style={box, fill=yellow!22, align=left, font=\scriptsize, text width=44mm, inner xsep=6pt, inner ysep=4pt},
  opbox/.style={box, fill=orange!22, align=left, font=\scriptsize, text width=52mm, inner xsep=6pt, inner ysep=4pt},
  beliefbox/.style={box, fill=gray!15, font=\footnotesize, align=center, minimum width=30mm},
  miniarr/.style={->, thick, shorten <=1pt, shorten >=1pt},
  edgelabel/.style={font=\scriptsize, fill=white, inner sep=1pt, align=center},
  rowlabel/.style={font=\bfseries\footnotesize, anchor=east},
]

\node[rowlabel] (toplabel) at (0, 0) {Classical};

\node[vlmbox, right=10mm of toplabel] (cvlm) {Foundation\\model};
\node[geombox, below=10mm of cvlm] (cgeom) {Geometric\\channel};

\node[beliefbox, minimum height=24mm, minimum width=32mm, anchor=west]
  (cmap) at ([xshift=72mm]cvlm.east |- 0,-1) {Belief / Map};

\draw[miniarr] (cvlm.east) -- node[edgelabel, above]
  {raw $(\Conf, \phi_f)$ at $v^*$} (cvlm.east -| cmap.west);
\draw[miniarr] (cgeom.east) -- node[edgelabel, above]
  {assertion $\phi_g$ at $(v,t)$} (cgeom.east -| cmap.west);

\node[font=\small\itshape, gray!50!black, align=center, anchor=north]
  at ([yshift=-3mm]cmap.south)
  {each channel committed as-is within its own scope; no cross-channel reconciliation};

\node[rowlabel] (botlabel) at (0, -5.2) {Proposed};

\node[vlmbox, right=10mm of botlabel] (pvlm) {Foundation\\model};
\node[geombox, below=28mm of pvlm] (pgeom) {Geometric\\channel};

\node[calbox, right=20mm of pvlm, minimum height=18mm, anchor=west] (pcal) {%
  \textbf{Calibration}\\[1pt]
  reported $\Conf$ recalibrated against\\
  verified outcomes per context $\kappa$\\[2pt]
  conformal $\hat q_{1-\alpha}$ for prediction sets\\
  output: calibrated LR $\CalLR(e)$
};

\node[calbox, anchor=west, minimum height=22mm] (pcomp) at (pcal.west |- pgeom.center) {%
  \textbf{Compatibility}\\[1pt]
  $\mathrm{compat}(\phi_f, \phi_g) \in \{\top, \bot, \varnothing\}$\\[1pt]
  corroborates~($\top$): $\phi_g$ matches $\phi_f$\\
  contradicts~($\bot$): $\phi_g$ denies $\phi_f$\\
  channel deficit~($\varnothing$): $\phi_g\!=\!\bot_{\!\mathcal{G}}$\\
  \hspace*{2mm}\textit{(OOV, unseen, or unknown)}
};

\node[opbox, anchor=west, minimum height=42mm] (pupd) at ([xshift=10mm]pcal.east |- 0,-7.1) {%
  \textbf{Update operator}\\[2pt]
  if $\mathrm{compat}\!=\!\bot$:\\
  \hphantom{xx}emit $\bot$, enqueue $\mathcal{Q}_\bot$\\
  \hphantom{xx}refuse commit\\[3pt]
  else, candidate $\mathrm{Beta}(\alpha', \beta')$:\\
  \hphantom{xx}$\alpha' = \alpha + w T(v) \dfrac{\CalLR}{1+\CalLR}$\\
  \hphantom{xx}$\beta' = \beta + w T(v) \dfrac{1}{1+\CalLR}$\\
  \hphantom{xx}with $w = w_0\, \hat\kappa(e)$\\[3pt]
  revision: $D_{\mathrm{KL}}(\Belief\,\|\,\Belief') > \tau_{\mathrm{rev}}$?\\
  \hphantom{xx}yes $\to$ quarantine $\mathcal{Q}$\\
  \hphantom{xx}no $\to$ commit, append $\Prov$
};

\node[beliefbox, right=8mm of pupd, minimum height=24mm, minimum width=32mm] (pmap) {%
  \textbf{Belief / Map}\\[2pt]
  $\Belief(v,\phi)\!=\!\mathrm{Beta}(\alpha,\beta)$\\[2pt]
  mean $\dfrac{\alpha}{\alpha+\beta}$\\[1pt]
  ESS $\alpha+\beta$\\
  pred set $C_{1-\alpha}$
};

\node[box, fill=red!10, font=\scriptsize, align=center, anchor=north,
      minimum width=36mm, minimum height=12mm]
      (pq) at ([xshift=-22mm, yshift=-14mm]pupd.south)
      {\textbf{Revision quarantine $\mathcal{Q}$}\\[1pt]
       \itshape release on corroborate /\\\itshape timeout-$w/k$ / confirm};

\node[box, fill=red!22, font=\scriptsize, align=center, anchor=north,
      minimum width=36mm, minimum height=12mm]
      (pqbot) at ([xshift=22mm, yshift=-14mm]pupd.south)
      {\textbf{Disagreement $\mathcal{Q}_\bot$}\\[1pt]
       \itshape inconsistency signal;\\\itshape never committed};

\draw[dashed, gray!60, thick]
  ([xshift=-5mm, yshift=-38mm]toplabel.west) --
  ([yshift=-38mm]toplabel.west -| pmap.east);

\draw[miniarr] (pvlm.east) -- node[edgelabel, above, align=center]
  {evidence event\\$e\!=\!(T,\phi_f,\Conf,\kappa)$} (pcal.west);

\draw[miniarr] (pgeom.east) -- node[edgelabel, above]
  {$\phi_g$ or (no claim)} (pcomp.west);

\draw[miniarr] (pcal.east) -- node[edgelabel, above, align=center]
  {$\CalLR(e),\,\hat\kappa(e)$} (pcal.east -| pupd.west);

\draw[miniarr] (pcomp.east) -- node[edgelabel, above]
  {$\mathrm{compat}$} (pcomp.east -| pupd.west);

\draw[miniarr] (pupd.east) -- node[edgelabel, above]
  {$\Belief'(v,\phi)$} (pmap.west);

\draw[->, thick, dashed, gray!75]
  ([xshift=8mm]pcomp.north) -- ([xshift=8mm]pcal.south);
\node[font=\small, text=gray!85!black, fill=white, inner sep=2.5pt, align=center, draw=gray!40, rounded corners=1.5pt, dashed]
  at ($([xshift=8mm]pcomp.north)!0.5!([xshift=8mm]pcal.south)$)
  {\itshape verification: append $(\Conf, y)$ to\\\itshape calibration buffer when $\phi_g$ decisive};

\draw[miniarr] ([xshift=-25mm]pupd.south) -- ([xshift=-25mm]pupd.south |- pq.north)
  node[edgelabel, midway, align=center]
  {if revision\\$D_{\mathrm{KL}}\!>\!\tau_{\mathrm{rev}}$};
\draw[miniarr] ([xshift=-13mm]pupd.south |- pq.north) -- ([xshift=-13mm]pupd.south)
  node[edgelabel, midway] {release};

\draw[miniarr] ([xshift=22mm]pupd.south) -- ([xshift=22mm]pupd.south |- pqbot.north)
  node[edgelabel, midway, align=center]
  {if $\mathrm{compat}\!=\!\bot$\\emit $\bot$};

\end{tikzpicture}%
}
\caption{\textbf{Evidence-fusion pipelines.} Top: classical (each channel commits in its own scope; the geometric channel's assertion never reaches the foundation-model commit gate). Bottom: proposed operator with calibration $\CalLR$, compatibility verdict $\top$/$\bot$/$\varnothing$, revision quarantine $\mathcal{Q}$, and disagreement sink $\mathcal{Q}_\bot$; verification (dashed) feeds the calibration buffer when $\phi_g$ is decisive. Symbols defined in Defs.~\ref{def:map}--\ref{def:update}.}
\label{fig:pipeline_comparison}
\end{figure}
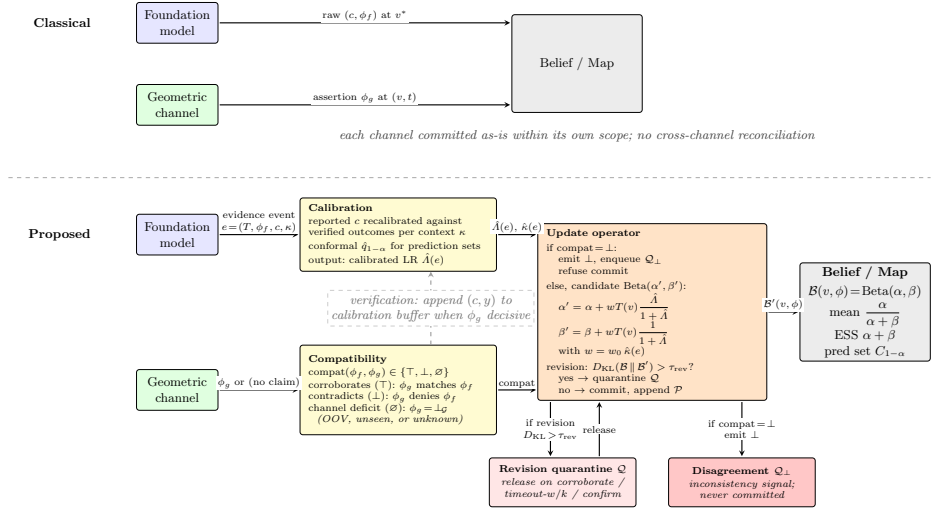

\subsection{Map and Evidence Types}
\label{sec:formaldef}

The definitions below formalize how foundation-model claims are represented as uncertain evidence over the map structure introduced informally above.

\begin{definition}[Spatial-semantic map]
\label{def:map}
A spatial-semantic map is a tuple $\Map = (\Geom, \Pred, \Belief)$, where $\Geom = (V, E)$ is a graph whose vertices $V$ represent \emph{map elements} (geometric primitives such as voxels or pose-graph nodes, or higher-level topological-semantic entities such as scene-graph nodes for objects, rooms, or regions), and edges $E$ represent geometric or topological relations. $\Pred$ is a space of \emph{predicates} that may be asserted of map elements, treated as encodable claims (typically VLM or LLM feature representations) without commitment to a fixed vocabulary. $\Belief : V \times \Pred \to \mathcal{D}$ is a \emph{belief function} that returns one Beta distribution per (element, predicate) pair; we instantiate $\mathcal{D}$ as $\Belief(v, \phi) = \mathrm{Beta}(\alpha_{v, \phi}, \beta_{v, \phi})$ so the subscripts make explicit that each pair carries its own parameters.
\end{definition}

We treat $V$ as heterogeneous because different scene-graph systems make different commitments and the framework should accommodate them.

\begin{definition}[Foundation-model evidence event]
\label{def:evidence}
An evidence event is a tuple $e = (T, \phi, \Conf, \kappa)$, where $T : V \to [0, 1]$ with $\sum_{v \in V} T(v) \leq 1$ is the \emph{targeting distribution} over map elements; $\phi \in \Pred$ is the asserted predicate; $\Conf \in [0, 1]$ is the raw confidence; and $\kappa \in K$ is a \emph{context tag} (source-model identity, prompt version, modality) that the operator uses to partition its calibration buffer.
$\kappa$ is opaque to the operator; the listed examples are representative, but in general $\kappa$ should include any decoding-time hyperparameter that perturbs the foundation model's confidence distribution.
We refer to $T$ as the targeting distribution in that the evidence is targeting an inference over part of the map, but may or may not be accepted.
In the simplest case $T$ concentrates on a single element $v^*$; the distribution form handles cases where the foundation model's grounding (e.g., a projected bounding box) overlaps multiple map elements.
\end{definition}

\paragraph{Constructing $T$.}
The targeting distribution is built by the \emph{evidence extractor}, the wrapper around the foundation-model query that converts its output into the evidence-event tuple (Figure~\ref{fig:pipeline_comparison}).
One natural construction, used in our experiments, is projected-bounding-box intersection-over-union between the foundation-model query region and detected map elements; other schemes (semantic similarity over learned embeddings, user-supplied association) can be substituted without affecting the operator.
Mass below a small threshold $\tau_T$ is discarded, which handles the targeting-hallucination case where the foundation model refers to an object that has no projection onto any map element.

\begin{definition}[Update operator]
\label{def:update}
An update operator is a function $\Upd : \Map^* \times \Ev \to \Map^*$, where $\Map^*$ is the space of valid maps and $\Ev$ is the space of evidence events. We write $\Map_e = \Upd(\Map, e)$.
\end{definition}

\subsection{Soundness Properties}
\label{sec:properties}

A sound foundation-model-fused update operator must satisfy structural and quantitative invariants beyond the two we focus on below; we reserve a full enumeration for an expanded work.\arxivonly{ Closure, bounded success, internal consistency under coherent evidence, $\eta$-stability, and $\delta$-conservativity are stated formally as Properties~\ref{prop:closure}--\ref{prop:conservativity} in \S\ref{sec:supp_properties}; rates for our operator ($\eta = O(1/(\alpha + \beta))$; $\delta = 0$ for non-revision updates within the un-compacted horizon, $\delta = O(1/\kappa_{\min}^2)$ for revision updates) are derived in \S\ref{sec:supp_proofs}.}

The two soundness properties the framework's composition addresses, and that current persistent-mapping systems most consistently violate, are population-level miscalibration on the foundation-model channel and silent per-event disagreement between the foundation-model and geometric channels.
We state both below.
Two auxiliary objects appear: the \emph{calibratable regime} $\CRgm \subseteq V \times \Pred$, the set of element-predicate pairs for which the geometric channel supplies verifiable outcomes and the calibration buffer has accumulated at least one verification event for the $(\phi, \kappa)$ context (Section~\ref{sec:asymmetry}); and the \emph{compatibility predicate} $\mathrm{compat}(\phi_f, \phi_g) \in \{\top, \bot, \varnothing\}$, where $\top$ denotes that the geometric channel's assertion $\phi_g$ at $(v, t)$ corroborates the foundation-model assertion $\phi_f$, $\bot$ that it contradicts, and $\varnothing$ that the geometric channel has no verifiable assertion over $\phi_f$.

\begin{property}[$(1-\alpha)$-coverage on the calibratable regime]
\label{prop:coverage}
For $(v, \phi) \in \CRgm$, the operator emits a $(1-\alpha)$-prediction set over the binary outcome $\{0, 1\}$ that contains the ground-truth outcome $y^*_{v, \phi}$ with marginal probability at least $1-\alpha$; outside $\CRgm$, the operator signals that no coverage claim applies. \emph{This property rules out prediction sets that silently fail to contain ground truth on the calibratable regime, and spurious coverage claims off it.} In plain terms: on predicates the robot can verify, the operator knows what it does not know.
\end{property}

\begin{property}[Cross-channel disagreement detectability]
\label{prop:comod}
If $\mathrm{compat}(\phi_f, \phi_g) = \bot$ at the moment of event $e_f$ targeting $v$, the operator emits $\bot$ and does not commit $e_f$ at $(v, \phi_f)$. \emph{This property rules out silent per-event disagreement passing into the belief.} In plain terms: a foundation-model claim that contradicts the geometric channel at the moment it is made cannot enter the belief.
\end{property}

Properties~\ref{prop:coverage} and~\ref{prop:comod} correspond to the population-level and per-event symptoms of Section~\ref{sec:intro}; their soundness is the load-bearing claim of the operator's calibration and compatibility-check stages, respectively.

Property~\ref{prop:comod} is intentionally asymmetric, privileging the geometric channel as the verification source.\arxivonly{ Order-dependence and the $\{0,1\}$ coverage formulation are discussed in \S\ref{sec:supp_property_commentary}.}
\section{Calibrated Revision-Aware Update Operator}
\label{sec:method}

The operator combines two mechanisms applied in series to every foundation-model event: a \emph{per-class calibrated commit gate} (\S\ref{sec:method:calibration}) that reweights the event by the foundation model's empirical reliability on its asserted class, and a \emph{per-event conflict-drop window} (\S\ref{sec:method:comod}) that consults the geometric channel and refuses commit when the two channels contradict each other.
\S~\ref{sec:method:revision}--\ref{sec:method:coverage} add a revision-detection quarantine for large posterior changes, a provenance log that makes any commit reversible, and the coverage guarantee Property~\ref{prop:coverage} establishes.

\subsection{Per-Class Calibrated Commit Gate}
\label{sec:method:calibration}

Foundation models do not come with characterized error distributions, so any operator that produces certified beliefs must include a calibration component.
We minimize the operational burden by using the robot's own perception as the calibration source and maintaining calibration online from streaming data.
Calibration points accumulate at zero marginal cost whenever the foundation model asserts a predicate the geometric stack can verify; hundreds to thousands accumulate over a typical deployment without dedicated annotation.

For an evidence event $e = (T, \phi, \Conf, \kappa)$ with $T$ concentrated on $v^*$, the operator maintains a streaming buffer $\mathcal{C} = \{(e_i, y_i)\}_{i=1}^n$ of past evidence-outcome pairs and uses it to produce two complementary calibrators.
The nonconformity score\begin{equation}
s(e, y) = -y \log \Conf - (1 - y) \log(1 - \Conf)
\label{eq:nonconformity}
\end{equation}is the negative log-likelihood the foundation model assigns to $y$ given its reported confidence; its empirical $(1 - \alpha)$-quantile $\hat{q}_{1-\alpha}$ of $\{s(e_i, y_i)\}_{i=1}^n$ \cite{angelopoulos2023gentle}, fit per context $\kappa$ and predicate $\phi$, defines the conformal prediction-set threshold that drives the coverage guarantee (Prop.~\ref{prop:coveragethm}).
Separately, we fit a monotone calibrator $\hat{p}(y = 1 \mid \Conf, \kappa, \phi)$ via isotonic regression on the same per-$(\kappa, \phi)$ slice of the buffer; the calibrated likelihood ratio used as the per-event commit weight is\begin{equation}
\CalLR(e) = \frac{\hat{p}(\Conf \mid y = 1, \kappa, \phi)}{\hat{p}(\Conf \mid y = 0, \kappa, \phi)},
\label{eq:lr}
\end{equation}with the conditional densities recovered from the isotonic calibrator and the empirical class base rate.
Stratifying the calibrator per predicate $\phi$ (a class-stratified Mondrian conformal regime \cite{vovk2005algorithmic}) is what makes the gate \emph{per class}: a foundation model reliable on one class but not another picks up different $\CalLR$ weights and crosses the commit threshold at different posterior strengths.
Sparsely-populated $(\phi, \kappa)$ buckets shrink toward the global estimate, with shrinkage controlled by a calibration-quality multiplier $\hat{\kappa}(e) \in (0, 1]$ defined as the matching context's effective sample size divided by a target threshold.

\subsection{Per-Event Commit Gate and Conflict-Drop Window}
\label{sec:method:comod}

The compatibility predicate consults the geometric channel at the moment of the event and returns a three-valued verdict $\mathrm{compat}(\phi_f, \phi_g) \in \{\top, \bot, \varnothing\}$.
Concretely, our implementation derives the verdict from a per-predicate subsumption table: $\top$ if the geometric channel's class label $\phi_g$ at $(v,t)$ lies in $\phi_f$'s positive subsumption set (e.g., \texttt{is\_car} subsumes \{car, truck, bus\}), $\bot$ if $\phi_g$ lies in $\phi_f$'s negative subsumption set (a class that explicitly disagrees, e.g., \texttt{is\_pedestrian} asserted on a panoptic-car region), and $\varnothing$ if $\phi_g$ is out of vocabulary for $\phi_f$ or the geometric channel cannot assert at $(v,t)$.

On $\top$ or $\varnothing$, the operator applies the calibrated update of \S\ref{sec:method:calibration} unchanged.
On $\bot$, the operator emits the inconsistency symbol, routes the pair to a \emph{disagreement queue} $\mathcal{Q}_\bot$, and does \emph{not} commit the event.
Queued disagreements resolve by one of three conditions: later $\top$ corroboration at the same $(v, \phi_f)$ within the conflict-drop window $T_{\mathrm{disagree}}$ (marked resolved but \emph{not} retroactively committed; committing it would erase the disagreement signal); timeout (released unresolved); or explicit operator confirmation.

This per-event refusal is the only mechanism that prevents a $\bot$-marked event from entering the posterior at all; routine noisy-but-plausible commits, including a possibly-spurious first detection, are buffered by three downstream properties.
First, the Beta posterior is initialized to the uniform prior $\mathrm{Beta}(1, 1)$, so accumulated evidence must overcome the prior before the commit threshold $\tau$ fires.
Second, the per-event update (Eq.~\eqref{eq:betaupdate} below) is fractional in $\CalLR$: a single event with $\CalLR \approx 5$ shifts the posterior mean by roughly $0.1$, not toward certainty.
Third, any single update that flips the posterior mode is intercepted by the revision-detection step (\S\ref{sec:method:revision}) and quarantined rather than committed immediately.

\subsection{Revision Detection, Quarantine, and Retraction}
\label{sec:method:revision}
\label{sec:provenance}

After the compatibility check passes, the operator applies a fractional pseudo-count update.
Given $\CalLR(e)$ and current belief $\Belief(v, \phi) = \mathrm{Beta}(\alpha, \beta)$, the candidate updated belief is $\mathrm{Beta}(\alpha', \beta')$ with\begin{equation}
\alpha' = \alpha + w T(v) \frac{\CalLR(e)}{1 + \CalLR(e)}, \quad \beta' = \beta + w T(v) \frac{1}{1 + \CalLR(e)},
\label{eq:betaupdate}
\end{equation}where $w = w_0 \hat{\kappa}(e)$ is the per-evidence weight modulated by the calibration-quality multiplier; this is a Bayesian odds update, $\alpha'/\beta' = (\alpha/\beta) \cdot \CalLR(e)$ in expectation.
For general targeting, Eq.~\eqref{eq:betaupdate} is applied to each $v$ with $T(v) > \tau_T$, with $T(v)$ as multiplicative weight; mass below $\tau_T$ is discarded.

The candidate $\Belief'$ is committed only if it does not represent a \emph{revision}, defined by the divergence test $D_{\mathrm{KL}}(\Belief ~\|~ \Belief') > \tau_{\mathrm{rev}}$.
Revision events enter the \emph{revision quarantine queue} $\mathcal{Q}$ (distinct from $\mathcal{Q}_\bot$) and commit when corroborating evidence arrives within $T_{\mathrm{cor}}$ steps, when the quarantine times out (committed at reduced weight $w/k$), or under operator confirmation; disconfirming evidence within the window discards the quarantined evidence.

Approximate retractability of any specific evidence event is required for safe deployment.
We implement this via an explicit provenance store $\Prov$ keyed to each element-predicate pair: for each $(v, \phi)$, an ordered log of triples $(e_t, \Delta_t^{\alpha}, \Delta_t^{\beta})$ records each updating event and its pseudo-count increments, with current belief reconstructed as\begin{equation}
(\alpha_{v, \phi}, \beta_{v, \phi}) = (\alpha_0, \beta_0) + \sum_{t \in \Prov(v, \phi)} (\Delta_t^{\alpha}, \Delta_t^{\beta}).
\label{eq:provrecon}
\end{equation}
The retraction operator $\Retr$ removes a specific event by recomputing the belief without its increments; this is exact for non-revision updates; for revision updates committed via corroboration, retraction also removes dependent corroborating evidence, with cascading retraction bounded by $T_{\mathrm{cor}}$.
Evidence older than horizon $T_{\mathrm{prov}}$ is compacted into a single equivalent increment.\arxivonly{ Retraction triggers, the compaction protocol, and hyperparameter rationale are detailed in \S\ref{sec:supp_operator}.}

Algorithm~\ref{alg:comod} ties the three mechanisms together for the full co-modulated update $\Upd_{\mathrm{comod}}$.
Per-evidence cost is one geometric-channel query, one compatibility evaluation, one isotonic-regression query (logarithmic in matching-context buffer size), and a constant-time Beta update; the disagreement queue is bounded by $T_{\mathrm{disagree}}$ times the per-step disagreement rate, which is empirically small.\arxivonly{ The calibration-only variant $\Upd_{\mathrm{cal}}$ that omits the per-event compatibility check is stated as Algorithm~\ref{alg:cal-update} in \S\ref{sec:supp_algorithms}.}

\begin{algorithm}[!ht]
\caption{Co-modulated calibrated update $\Upd_{\mathrm{comod}}$}
\label{alg:comod}
\begin{algorithmic}[1]
\Require Initial map $\Map_0$, calibration buffer $\mathcal{C}$, provenance store $\Prov$; hyperparameters $\alpha, \tau_{\mathrm{rev}}, T_{\mathrm{cor}}, T_{\mathrm{disagree}}, k, w_0, T_{\mathrm{prov}}$; geometric channel state provider $\Gamma$
\State Initialize quarantine queue $\mathcal{Q}\!\gets\!\emptyset$, disagreement queue $\mathcal{Q}_\bot\!\gets\!\emptyset$
\For{each incoming evidence $e_t = (T, \phi_f, \Conf, \kappa)$ at time $t$}
    \For{each $v \in V$ with $T(v) > \tau_T$}
        \State $\phi_g \gets \Gamma(v, \phi_f, t)$ \Comment{geometric-channel assertion at $(v,t)$}
        \State $\mathrm{compat} \gets \mathrm{compat}(\phi_f, \phi_g)$ with sign from $\Conf \geq 0.5$
        \If{$\mathrm{compat} = \bot$} \Comment{per-event refusal (Prop.~\ref{prop:comod})}
            \State Enqueue $(e_t, v, \phi_f, \phi_g)$ in $\mathcal{Q}_\bot$; append $\bot$ to $\Prov(v, \phi_f)$
            \State \textbf{continue}
        \EndIf
        \State $\CalLR \gets$ \Call{IsotonicLR}{$\Conf, \kappa, \phi_f, \mathcal{C}$} stratified by $(\kappa, \phi_f)$
        \State $\hat{\kappa}(e_t) \gets$ multiplier from local calibration density
        \State Compute $(\alpha', \beta')$ via Eq.~\eqref{eq:betaupdate} with $w = w_0\,\hat{\kappa}(e_t)$
        \State $D \gets D_{\mathrm{KL}}(\mathrm{Beta}(\alpha, \beta) \,\|\, \mathrm{Beta}(\alpha', \beta'))$
        \If{$D \leq \tau_{\mathrm{rev}}$} \Comment{non-revision: commit}
            \State $(\alpha, \beta) \gets (\alpha', \beta')$; append to $\Prov(v, \phi_f)$
        \Else \Comment{revision: quarantine}
            \State Enqueue $(e_t, v, \phi_f)$ in $\mathcal{Q}$; \Call{ProcessQuarantine}{$\mathcal{Q}, t$}
        \EndIf
        \If{$\mathrm{compat} = \top$}
            \State \Call{ResolveByLaterAgreement}{$\mathcal{Q}_\bot, (v, \phi_f), t$}
        \EndIf
    \EndFor
    \For{each calibration outcome $(e_\tau, y_\tau)$ available at $t$}
        \State Append $(e_\tau, y_\tau)$ to $\mathcal{C}$; update adaptive threshold $\hat{q}_{1-\alpha}$
    \EndFor
    \State \Call{TickDisagreementQueue}{$\mathcal{Q}_\bot, t, T_{\mathrm{disagree}}$} \Comment{timeouts}
    \If{$t \bmod T_{\mathrm{prov}} = 0$} \Call{CompactProvenance}{$\Prov, t - T_{\mathrm{prov}}$} \EndIf
\EndFor
\end{algorithmic}
\end{algorithm}

\subsection{Coverage Guarantee}
\label{sec:method:coverage}

\begin{proposition}[Marginal coverage of $\Upd_{\mathrm{cal}}$]
\label{prop:coveragethm}
Let $\mathcal{C}$ be the calibration buffer at the time of evidence $e$, and assume the calibration data $\{(e_i, y_i)\}_{i=1}^n \subset \mathcal{C}$ matching the context $\kappa$ are exchangeable with $(e, y)$, where $y$ is the eventual ground-truth outcome. Assume $\hat{\kappa}(e) \geq \kappa_{\min} > 0$. Then the prediction set $C_{1-\alpha}(v, \phi) \subseteq \{0, 1\}$ produced by $\Upd_{\mathrm{cal}}$ satisfies
\[
\PR\left( y^*_{v, \phi} \in C_{1-\alpha}(v, \phi) \right) \geq 1 - \alpha - \frac{1}{n + 1}.
\]
\end{proposition}

\arxivonly{\noindent Proof in \S\ref{sec:supp_proof_coverage}. }The $1/(n+1)$ term shrinks as the calibration buffer grows; sequential coverage over an update sequence is maintained online via the adaptive procedure of \cite{gibbs2021adaptive}.\arxivonly{ Companion propositions for stability ($\eta = O(1/(\alpha + \beta))$) and conservativity ($\delta = 0$ within the un-compacted horizon; $\delta = O(1/\kappa_{\min}^2)$ for revision-scale updates) are stated and proved in \S\ref{sec:supp_properties}.}

\section{Experimental Setup}
\label{sec:experiments}

We evaluate on two real-data environments: KITTI-360 \cite{liao2023kitti360} 
(5 sequences: \texttt{00}, \texttt{04}, \texttt{06}, \texttt{09}, \texttt{10}\footnote{All KITTI-360 sequence references use the shorthand \texttt{(XX)} for \texttt{drive\_00XX\_sync}.}; left RGB at $2$~Hz; per-pixel panoptic labels as geometric ground truth) for outdoor driving, and ScanNet \cite{dai2017scannet} ($50$ indoor scenes, ConceptGraphs \cite{gu2024conceptgraphs} backbone instrumented to expose the update operator as a swap-in component) for indoor navigation.
We use the datasets' ground-truth poses (no SLAM) and treat the geometric channel as deterministic at each query; the compatibility predicate's three-valued verdict ($\bot$/$\top$/$\varnothing$) is the operator's only handle on geometric-channel uncertainty.
KITTI bounding boxes come from YOLOv11s~\cite{jocher2023ultralytics}, held fixed across VLM substitutions and excluded from the $\kappa$ context.
We evaluate two foundation-model backbones, LLaVA-1.6 and Qwen3-VL ($\kappa$ held constant per deployment), on per-class object-finding, deployable-channel performance, belief preservation across weak-evidence transitions, and thing-class compositional retrieval.

\paragraph{Geometric channels.}
We compare an oracle channel (panoptic GT on KITTI, mesh-aggregation class labels on ScanNet) with a deployment channel (Mask2Former \cite{cheng2022mask2former}, mIoU $0.427$ on \texttt{00} and $0.449$ on \texttt{10} with no fine-tuning).
The comparison places the operator under an oracle-quality channel against a realistic deployment-grade one.

\paragraph{Predicate vocabulary and compound queries.}
Every compound retrieval query joins one VLM-typed predicate with one or more geometric-channel context predicates that supply spatial, dynamic, or visibility context.
For example, \emph{parked car near crossing} pairs the VLM-typed \texttt{is\_car} with geometric \texttt{is\_stationary}, \texttt{is\_near\_road\_px}, and $\neg$\texttt{is\_near\_pedestrian}; \emph{bookshelf facing seating area} pairs \texttt{is\_bookshelf} with \texttt{is\_facing\_seating} and \texttt{is\_accessible}.
The VLM-typed component is the calibration target; the geometric components are computed from the geometric channel plus dataset-level structure (3D bbox velocity, road graph, field-of-view tests) and treated as authoritative.
Each dataset has three ``original'' compounds plus a 2026-05 expansion that composes additional compounds from the same per-component beliefs at zero additional foundation-model cost.\arxivonly{ Full decompositions in Table~\ref{tab:queries}.}

\arxivonly{\begin{table}[t]
\centering\small
\renewcommand{\arraystretch}{1.2}
\setlength{\tabcolsep}{6pt}
\caption{Compound-query decompositions used in the evaluation. Each query is the conjunction of one VLM-typed (foundation-channel) predicate with one or more geometric-channel predicates. ``$\neg$'' denotes negation.}
\label{tab:queries}
\begin{tabular}{p{0.32\linewidth} p{0.18\linewidth} p{0.42\linewidth}}
\toprule
Compound query & VLM-typed & Geometric-channel context \\
\midrule
\multicolumn{3}{l}{\textit{KITTI-360 (original)}} \\
parked car near crossing      & \texttt{is\_car}          & \texttt{is\_stationary}, \texttt{is\_near\_road\_px}, $\neg$\texttt{is\_near\_pedestrian} \\
pedestrian approaching road   & \texttt{is\_pedestrian}   & \texttt{is\_on\_sidewalk}, \texttt{is\_near\_road\_px} \\
signaled intersection marker  & \texttt{is\_traffic\_sign}& \texttt{is\_at\_intersection}, \texttt{is\_visible\_in\_drive\_direction} \\
\multicolumn{3}{l}{\textit{KITTI-360 (expansion)}} \\
car at intersection           & \texttt{is\_car}          & \texttt{is\_at\_intersection} \\
car near pedestrian           & \texttt{is\_car}          & \texttt{is\_near\_pedestrian} \\
pedestrian at intersection    & \texttt{is\_pedestrian}   & \texttt{is\_at\_intersection} \\
pedestrian jaywalking         & \texttt{is\_pedestrian}   & $\neg$\texttt{is\_on\_sidewalk}, \texttt{is\_near\_road\_px} \\
moving car in traffic         & \texttt{is\_car}          & $\neg$\texttt{is\_stationary}, \texttt{is\_near\_road\_px} \\
\midrule
\multicolumn{3}{l}{\textit{ScanNet (original)}} \\
chair near window             & \texttt{is\_chair}        & \texttt{is\_near\_window} \\
bookshelf facing seating area & \texttt{is\_bookshelf}    & \texttt{is\_facing\_seating}, \texttt{is\_accessible} \\
unobstructed workspace surface& \texttt{is\_table\_or\_desk}& \texttt{is\_workspace\_height}, \texttt{is\_unobstructed} \\
\multicolumn{3}{l}{\textit{ScanNet (expansion)}} \\
chair at workspace height     & \texttt{is\_chair}        & \texttt{is\_workspace\_height} \\
chair at workspace, unobstructed & \texttt{is\_chair}     & \texttt{is\_workspace\_height}, \texttt{is\_unobstructed} \\
accessible table              & \texttt{is\_table\_or\_desk}& \texttt{is\_accessible} \\
accessible bookshelf          & \texttt{is\_bookshelf}    & \texttt{is\_accessible} \\
\bottomrule
\end{tabular}
\end{table}}

\paragraph{Operators, hyperparameters, and metrics.}
Table~\ref{tab:setup} lists the six compared operators with their key hyperparameters and the per-claim metrics.

\begin{table}[h]
\centering\scriptsize
\renewcommand{\arraystretch}{1.1}
\setlength{\tabcolsep}{4pt}
\caption{Operators, hyperparameters, and metrics. The first four operators characterize current practice; $\Upd_{\mathrm{cal}}$ adds the calibrated commit gate (\S\ref{sec:method:calibration}); $\Upd_{\mathrm{comod}}$ extends $\Upd_{\mathrm{cal}}$ with the per-event compatibility check (\S\ref{sec:method:comod}). All operators are evaluated at commit threshold $\tau=0.5$.}
\label{tab:setup}
\begin{tabular}{p{0.20\linewidth} p{0.74\linewidth}}
\toprule
\multicolumn{2}{l}{\emph{Operators}} \\
Overwrite & sets $\Belief(v,\phi)$ to a high-confidence belief on each high-$\Conf$ event \\
Majority vote & accumulates votes weighted by raw $\Conf$, no calibration \\
Naive Bayesian & treats $\Conf$ as a calibrated likelihood with fixed assumed reliability \\
Refined baseline & temporal-majority windowed filter \\
$\Upd_{\mathrm{cal}}$ & calibrated commit gate only \\
$\Upd_{\mathrm{comod}}$ & calibrated commit gate $+$ per-event compatibility check \\
\midrule
\multicolumn{2}{l}{\emph{Hyperparameters for $\Upd_{\mathrm{cal}}$ / $\Upd_{\mathrm{comod}}$}} \\
& $\alpha = 0.1$,\; $\tau_{\mathrm{rev}} = 0.5$ nats,\; $T_{\mathrm{cor}} = 10$,\; $k = 4$,\; $w_0 = 1$,\; $T_{\mathrm{prov}} = 1000$,\; $T_{\mathrm{disagree}} = 10$ \\
\midrule
\multicolumn{2}{l}{\emph{Metrics}} \\
Per-class IoU & at $\tau = 0.5$ on the threshold-binarized scene graph \\
Compositional retrieval & F1 at $p \geq 0.5$ with class-conditional recalibration \\
Deployable channel & per-event $\bot$-emission rate and operator-final mean IoU under M2F vs.\ panoptic GT \\
Cross-VLM invariance & Brier score on the belief-preservation pool under VLM substitution \\
\bottomrule
\end{tabular}
\end{table}

\section{Results}
\label{sec:results}

We make four empirical claims about the map the operator produces.
The first three concern the map's content (per-class IoU, compositional retrieval quality, deployability); the fourth concerns robustness under a foundation-model swap.

\paragraph{Per-class IoU.}
We report per-class IoU at commit threshold $\tau=0.5$ as the headline map metric.
In our view this is the most defensible single number for a committed map: downstream actions (retrieval, planning, task-conditioned querying) operate on the per-class binarized commitments, and IoU penalizes both over-commitment (low precision under overwrite, majority, naive-Bayes) and refused-rare-class collapse (the refined baseline's $0.010$ IoU on \texttt{is\_traffic\_sign}) on a single axis.
Our operator $\Upd_{\mathrm{comod}}$ wins every per-class IoU column on both datasets, by margins of $0.04$--$0.47$ on KITTI-360 and $0.42$--$0.77$ on ScanNet over the next-best operator (Table~\ref{tab:miou}).
The cleanest single comparison is dominant-class precision on KITTI: $\Upd_{\mathrm{comod}}$ commits \texttt{is\_car} at $99.7\%$ precision against $\Upd_{\mathrm{cal}}$'s $43.9\%$ on the same event pool.
$\Upd_{\mathrm{comod}}$'s gain comes from a per-event refusal that adapts to where the foundation-model channel is unreliable, beyond what calibration's sharpening or shrinkage delivers on its own.

\begin{table}[!htbp]
\centering\scriptsize
\renewcommand{\arraystretch}{1.15}
\setlength{\tabcolsep}{3.5pt}
\caption{Per-class IoU $\uparrow$ at commit threshold $0.5$ at the end of trajectory, pooled across $5$ KITTI-360 sequences and $50$ ScanNet scenes. KITTI uses panoptic GT as truth; ScanNet uses mesh-aggregation class labels. ScanNet's candidate-class filter at intake suppresses cross-class false positives by construction; IoU on ScanNet therefore collapses to recall on the candidate-class universe.}
\label{tab:miou}
\begin{tabular}{l cccc cccc}
\toprule
 & \multicolumn{4}{c}{KITTI-360} & \multicolumn{4}{c}{ScanNet} \\
\cmidrule(lr){2-5} \cmidrule(lr){6-9}
Operator & car & pedest. & sign & Mean & chair & bookshlf & tbl/desk & Mean \\
\midrule
Overwrite                       & 0.367 & 0.089 & 0.064 & 0.173 & 0.260 & 0.321 & 0.041 & 0.207 \\
Majority vote                   & 0.318 & 0.174 & 0.096 & 0.196 & 0.199 & 0.194 & 0.027 & 0.140 \\
Naive Bayesian                  & 0.318 & 0.174 & 0.096 & 0.196 & 0.199 & 0.194 & 0.027 & 0.140 \\
Refined baseline                & 0.603 & 0.091 & 0.010 & 0.235 & 0.000 & 0.000 & 0.000 & 0.000 \\
$\Upd_{\mathrm{cal}}$           & 0.331 & 0.120 & 0.089 & 0.180 & 0.464 & 0.452 & 0.120 & 0.345 \\
$\Upd_{\mathrm{comod}}$         & \textbf{0.796} & \textbf{0.211} & \textbf{0.560} & \textbf{0.522} & \textbf{0.879} & \textbf{0.885} & \textbf{0.889} & \textbf{0.884} \\
\bottomrule
\end{tabular}
\end{table}

\paragraph{Compositional retrieval.}
Table~\ref{tab:expanded} reports F1 on $15$ compound queries from KITTI-360 and ScanNet.
On dense-positive compounds (positive rate $10$--$60\%$) the gain is $0.04$--$0.35$ F1; the sparse-positive regime is sharper: when the positive rate falls below $1\%$, the cold-VLM commits indiscriminately (\emph{moving car in traffic}: $965$ commits, $5$ true positives, F1 $0.010$) while the operator commits at $5/5$ true positives (F1 $1.000$).
The expansion compounds illustrate the per-component-belief reuse property: \emph{car at intersection} and \emph{car near pedestrian} share \texttt{is\_car} with \emph{parked car near crossing}, so the operator pays VLM cost once and answers many compound questions; the cold-VLM baseline must re-prompt for each compound.
The same compounds with Qwen3-VL agree to within $0.015$ F1.

\begin{table}[!htbp]
\centering\scriptsize
\renewcommand{\arraystretch}{1.15}
\setlength{\tabcolsep}{4pt}
\caption{Compositional retrieval F1 $\uparrow$ at commit threshold $p \geq 0.5$, with class-conditional recalibration applied uniformly. \emph{Full pool} is the operator's complete retrieval set; \emph{intersection} is the visible-element subset the monolithic cold-VLM baseline is also defined on. Operator: $\Upd_{\mathrm{comod}}$. ``--'' marks intersections too small ($n < 20$) for stable 10-fold cross-validation.}
\label{tab:expanded}
\begin{tabular}{l rrr rrrr}
\toprule
                                 & \multicolumn{3}{c}{Full pool (operator)} & \multicolumn{4}{c}{Apples-to-apples intersection} \\
\cmidrule(lr){2-4} \cmidrule(lr){5-8}
                                 & $n$ & pos\,\% & F1 $\uparrow$ & $n$ & cold raw & cold $+$M & operator \\
\midrule
\multicolumn{8}{l}{\textit{KITTI-360, original compounds}} \\
parked car near crossing       & 3{,}910 & 58.1 & 0.746 & 2{,}309 & 0.706 & 0.826 & \textbf{0.950} \\
pedestrian approaching road    & 1{,}287 &  8.7 & 0.373 &    122 & 0.720 & 0.906 & \textbf{0.976} \\
signaled intersection marker   & 1{,}739 & 10.8 & 0.000 &    408 & 0.531 & 0.000 & \textbf{0.924} \\
\addlinespace[2pt]
\multicolumn{8}{l}{\textit{KITTI-360, expansion compounds}} \\
car at intersection            & 3{,}910 & 57.2 & 0.749 & 2{,}269 & 0.781 & 0.912 & \textbf{0.955} \\
car near pedestrian            & 3{,}910 &  6.0 & 0.000 & 2{,}263 & 0.144 & 0.000 & \textbf{0.601} \\
pedestrian at intersection     & 1{,}201 &  3.5 & 0.000 &     46 & 0.805 & 0.930 & \textbf{0.988} \\
pedestrian jaywalking          & 1{,}201 &  0.3 & 0.000 &     46 & 0.061 & 0.000 & \textbf{0.800} \\
moving car in traffic          & 3{,}910 &  0.1 & 0.000 & 2{,}295 & 0.010 & 0.000 & \textbf{1.000} \\
\midrule
\multicolumn{8}{l}{\textit{ScanNet, original compounds}} \\
chair near window              &     150 & 16.7 & 0.000 &      45 & 0.114 & 0.000 & 0.000 \\
bookshelf facing seating area  &      32 & 46.9 & 0.343 &       9 & 0.444 & --    & --    \\
unobstructed workspace surface &      77 & 14.3 & 0.000 &      36 & 0.143 & 0.000 & 0.000 \\
\addlinespace[2pt]
\multicolumn{8}{l}{\textit{ScanNet, expansion compounds}} \\
chair at workspace height      &     150 & 37.3 & 0.161 &      45 & 0.381 & 0.000 & \textbf{0.364} \\
chair at workspace, unobstructed & 150 & 11.3 & 0.000 &      45 & 0.343 & 0.000 & \textbf{0.400} \\
accessible table               &      77 & 59.7 & 0.752 &      36 & 0.512 & 0.455 & \textbf{0.800} \\
accessible bookshelf           &      32 & 65.6 & 0.769 &       9 & 0.600 & --    & --    \\
\bottomrule
\end{tabular}
\end{table}

\begin{figure}[!htbp]
\centering
\includegraphics[width=0.95\linewidth]{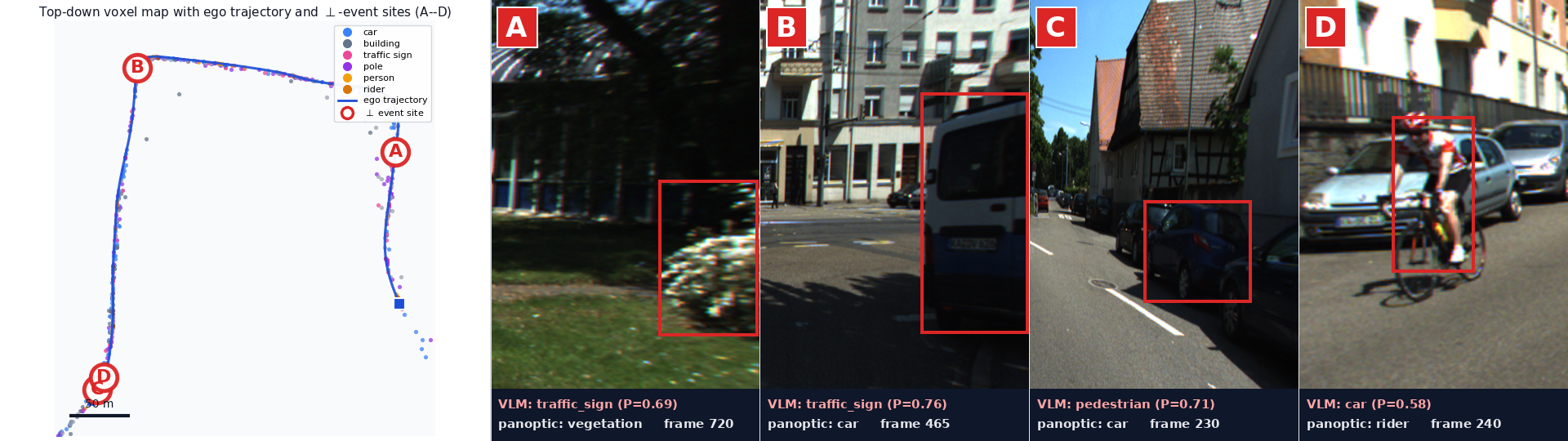}
\caption{$\Upd_{\mathrm{comod}}$ refusing foundation-model claims that contradict the geometric channel in a 60-second window of KITTI-360 \texttt{drive\_0010\_sync}. \emph{Left}: top-down voxel scatter colored by ground-truth class labels; ego trajectory in blue ($\blacktriangle$ start, $\blacksquare$ end); four red circles mark refused-claim sites \textbf{(A--D)}. \emph{Right}: image crops with the foundation-model query region; banners report the model's claim, its confidence, and the geometric channel's class assertion. The $1{,}255$ refusals in this window are claims a calibration-only operator and every evaluated baseline would have committed silently.}
\label{fig:voxblox_demo}
\end{figure}

\paragraph{Deployable channel.}
Swapping panoptic ground truth for Mask2Former as the geometric channel drops $\Upd_{\mathrm{comod}}$'s mean IoU from $0.522$ to $0.442$ (Table~\ref{tab:m2f_deploy}), localizing nearly all of that loss to \texttt{is\_traffic\_sign} recall ($\Delta$ IoU $-0.198$).
The per-event refusal rate matches within $1$pp across all three predicates, so the compatibility predicate's qualitative behavior survives the segmenter swap.
The remaining $0.442$ is still $1.9\times$ the IoU of any non-operator baseline in Table~\ref{tab:miou}: the framework gives up roughly a fifth of its lift to deployment realism and keeps the rest.
A voxel-backed instance on KITTI-360 \texttt{drive\_0010\_sync} (Figure~\ref{fig:voxblox_demo}) shows the operator refusing $1{,}255$ claims a calibration-only operator would have committed silently within a single $60$-second window.\arxivonly{ Extended pipeline description in \S\ref{sec:supp_voxel}.}

\begin{table}[!htbp]
\centering\scriptsize
\renewcommand{\arraystretch}{1.1}
\setlength{\tabcolsep}{5pt}
\caption{Deployable-channel substitution: $\Upd_{\mathrm{comod}}$'s threshold-binarized map quality on KITTI-360 ($5$ sequences, pooled) with $\phi_g$ = panoptic GT (oracle) versus $\phi_g$ = Mask2Former (deployment). Per-class IoU, precision, recall at commit threshold $\tau = 0.5$; $\Delta$ IoU is M2F $-$ GT.}
\label{tab:m2f_deploy}
\begin{tabular}{l ccc ccc c}
\toprule
 & \multicolumn{3}{c}{$\phi_g$ = panoptic GT} & \multicolumn{3}{c}{$\phi_g$ = Mask2Former} & \\
\cmidrule(lr){2-4} \cmidrule(lr){5-7}
Predicate & IoU $\uparrow$ & Prec $\uparrow$ & Rec $\uparrow$ & IoU $\uparrow$ & Prec $\uparrow$ & Rec $\uparrow$ & $\Delta$ IoU \\
\midrule
\texttt{is\_car}          & 0.796 & 0.997 & 0.798 & 0.784 & 0.983 & 0.795 & $-0.012$ \\
\texttt{is\_pedestrian}   & 0.211 & 0.212 & 0.980 & 0.180 & 0.187 & 0.822 & $-0.031$ \\
\texttt{is\_traffic\_sign}& 0.560 & 0.608 & 0.877 & 0.362 & 0.532 & 0.532 & $-0.198$ \\
\midrule
Mean                      & \textbf{0.522} & 0.606 & 0.885 & \textbf{0.442} & 0.567 & 0.717 & $-0.080$ \\
\bottomrule
\end{tabular}
\end{table}

\paragraph{Cross-VLM invariance.}
We test whether the operator's behavior depends on the specific foundation model by swapping LLaVA-1.6 for Qwen3-VL on a fixed evaluation pool of $982$ belief-preservation pairs (same map element, foundation model first gives a confident verdict then a weaker one later, the stress regime for any time-fusing operator, since the weak second observation can pull a correct strong-evidence belief in the wrong direction).
Under the VLM swap, $\Upd_{\mathrm{cal}}$'s Brier shifts $0.247 \to 0.243$ ($|\Delta| = 0.004$) and $\Upd_{\mathrm{comod}}$'s shifts $0.241 \to 0.240$ ($|\Delta| = 0.001$), while every uncalibrated baseline degrades by $0.06$--$0.11$ Brier (Overwrite $+0.109$, Majority vote $+0.079$, Naive Bayesian $+0.062$, Refined baseline $+0.069$).
When the deployed system swaps a VLM across software updates, the calibrated operators do not need re-tuning; the baselines effectively must be rebuilt.\arxivonly{ Property-verification measurements (coverage on the calibratable regime, belief-preservation Brier, pooled retrieval ECE, empirical conservativity) appear in supplementary Tables~\ref{tab:supp_coverage}, \ref{tab:supp_beliefpres}, \ref{tab:supp_retrieval_ece}, and \ref{tab:emp_conservativity_supp}.}

\section{Discussion}
\label{sec:discussion}

\paragraph{Where the framework applies.}
The framework is admissible whenever the geometric channel produces per-element assertions over a predicate vocabulary the foundation-model channel can also assert over, and whenever the map representation supports stable element identity over the operating horizon.
Concrete instantiations span scene-graph systems with class-labeled nodes, voxel grids with semantic labels, and instance-segmented panoptic maps.
The Mask2Former-substitution result demonstrates that the geometric channel need not be an oracle.

\paragraph{What the framework does not address.}
The compatibility predicate evaluates locally on a single $(v, \phi)$ pair and does not enforce conditional consistency across predicates (e.g., \texttt{is\_curb} as a function of \texttt{is\_road} and \texttt{is\_sidewalk} on neighboring elements).
The operator consumes foundation-model events as \emph{positive} assertions only; failure to observe an expected predicate (negative inference, e.g.\ ``no chair detected at this element across many queries, therefore probably no chair'') is not currently routed through the operator and is left to future work.
Hierarchical scene-graph substrates that carry temporal layers \cite{schmid2024khronos} or rich per-node open-vocabulary description layers \cite{gorlo2026daaam} introduce per-layer and cross-layer compatibility-checking requirements not addressed by the single-predicate, single-snapshot operator described here.
Map elements are assumed stable over the operating horizon; loop closure or pose-graph re-optimization that re-indexes voxels invalidates accumulated beliefs unless paired with belief migration.
Moving objects with identity tracking across frames are out of scope.
Each of these is a tractable extension within the same formalism.\arxivonly{ The supplementary material discusses these extensions individually.}

\paragraph{Choice of indoor benchmark.}
We evaluate on ScanNet v2 \cite{dai2017scannet} rather than ScanNet++ for ConceptGraphs-backbone compatibility and for the larger per-task pool ($1{,}513$ scenes vs $\sim$$460$) the headline retrieval and coverage results need for statistical power.
ScanNet++'s sub-centimeter Faro geometry and long-tail open-vocabulary labels are the natural follow-up to stress-test the per-event compatibility check at higher geometric fidelity and to widen the calibratable regime $\CRgm$ beyond the few-class taxonomy ScanNet v2 supports.

\paragraph{Where Property~\ref{prop:coverage} is bounded by data, not by assumption.}
The $(1-\alpha)$ guarantee is established under exchangeability in the long-run regime; in practice it is bounded below by the calibration buffer's effective sample size on the $(\phi, \kappa)$ context being queried.
Tasks that give the operator many observations per instance (object-finding while the robot drives through the scene, retrieval with many candidates per compound query) achieve nominal coverage; the belief-preservation task with one strong and one weak observation per instance does not.
The shortfall is stable across both VLMs we tested (Section~\ref{sec:results}, cross-VLM invariance result), confirming it is a property of the per-instance evidence budget rather than of the foundation model.
The adaptive conformal procedure of \cite{gibbs2021adaptive} maintains long-run marginal coverage under bounded distribution shift, which subsumes the temporal correlation present in real sensor streams.

\paragraph{Summary.}
The bounds the framework offers are properties of the per-instance evidence budget, not of either foundation model.
The refusal-and-calibration composition delivers substantially more accurate maps ($+0.34$ mean IoU on KITTI, $+0.54$ on ScanNet over the next-best calibration-only operator), more compositional true positives at higher precision than a monolithic-prompt baseline, deployment-quality behavior under an off-the-shelf segmenter, and invariance to foundation-model substitution.

%
%
\makeatletter
\renewenvironment{thebibliography}[1]
     {\par\addvspace{0.5\baselineskip}%
      \noindent{\bfseries\refname}\par\nopagebreak\vspace{0.3ex}%
      \scriptsize
      \list{\@biblabel{\@arabic\c@enumiv}}%
           {\settowidth\labelwidth{\@biblabel{#1}}%
            \leftmargin\labelwidth
            \advance\leftmargin\labelsep
            \setlength{\itemsep}{0pt plus 0.1ex}%
            \setlength{\parsep}{0pt}%
            \usecounter{enumiv}%
            \let\p@enumiv\@empty
            \renewcommand\theenumiv{\@arabic\c@enumiv}}%
      \sloppy\clubpenalty10000\widowpenalty10000
      \sfcode`\.\@m}
     {\def\@noitemerr
       {\@latex@warning{Empty `thebibliography' environment}}%
      \endlist}
\makeatother
\bibliographystyle{splncs04}
\bibliography{references}

\clearpage


\ifarxiv
  \input{supp/main}
\fi

\end{document}